\documentclass[11pt]{article}

\usepackage[final]{acl}

\usepackage{times}
\usepackage{latexsym}

\usepackage{url}            
\usepackage{booktabs}       
\usepackage{amsfonts}       
\usepackage{nicefrac}       
\usepackage{microtype}      
\usepackage[table]{xcolor}   
\usepackage{latexsym}
\usepackage{amssymb}
\usepackage{amsmath}
\usepackage{booktabs}
\usepackage{enumerate}
\usepackage{graphicx}
\usepackage{subfigure}
\usepackage{xspace}
\usepackage{float}
\usepackage{bbm}
\usepackage{bm}
\usepackage{multirow}
\usepackage{booktabs}
\usepackage{color}
\usepackage{framed}
\usepackage{stfloats}
\usepackage{iitem}
\usepackage{amssymb}
\usepackage{pifont}
\usepackage{arydshln}
\usepackage{enumitem}
\usepackage{wrapfig}
\usepackage{algorithm}
\usepackage{algorithmic}
\usepackage{array}
\usepackage{tabularx}
\usepackage{booktabs}
\usepackage{float}
\usepackage{cleveref}
\usepackage{colortbl}
\usepackage{etoolbox}
\usepackage{tcolorbox}
\usepackage{amsthm}
\usepackage{mathtools}

\theoremstyle{plain}

\theoremstyle{definition}

\theoremstyle{remark}

\usepackage[T1]{fontenc}

\usepackage[utf8]{inputenc}

\usepackage{microtype}

\usepackage{inconsolata}

\usepackage{graphicx}

\newcommand{\method}{{DIAG}\xspace}
\title{DIAG: \underline{D}iagnostic \underline{I}terative \underline{A}lignment and \underline{G}eneration for \\ Data-Efficient Mathematical Preference Distillation}

\author{
 \textbf{Guhan Chen\textsuperscript{1}\thanks{Equal Contribution.},~} 
 \textbf{Songtao Tian\textsuperscript{1$\ast$},~} 
 \textbf{Bohan Li\textsuperscript{2},~} 
 \textbf{Hejin Wang\textsuperscript{1},~} 
 \textbf{Yexin Xie\textsuperscript{3},~} 
 \textbf{Zixiong Yu\textsuperscript{1,3}\thanks{Corresponding Author.}}
\\
 \textsuperscript{1}Tsinghua University\quad
 \textsuperscript{2}Kyoto University\quad
 \textsuperscript{3}Nanjing University
\\
\texttt{chen-gh23@mails.tsinghua.edu.cn}\quad\texttt{xieye-xin@qq.com}\\
\texttt{\{tiansongtao.2020,libh19,wanghj20,yuzx19\}@tsinghua.org.cn}
}

\begin{document}
\maketitle
\begin{abstract}
Iterative preference optimization is essential for aligning Large Language Models on mathematical reasoning tasks, yet its efficiency is often throttled by signal scarcity: as the model improves, static problem sets become increasingly mismatched to the model's evolving competence, producing rollouts that are either too easy or too hard and therefore non-informative, which leads to a scarcity of valid preference pairs. We propose \method, a \textit{Diagnostic Iterative Alignment and Generation} framework that adaptively reshapes the practice distribution to increase informative supervision and focus training near the student's current competence boundary. \method consists of two phases: (1) \textit{diagnosing} valid preference-pair yield to calibrate the exploration--exploitation trade-off and allocate topic quotas via an Empirical Bayes shrinkage estimator, thereby prioritizing high-yield concepts; and (2) \textit{generating} targeted practice, where a teacher synthesizes variants from the student's failure traces.
We further provide a theoretical view interpreting \method as a teacher-mediated approximation to KL-regularized reweighting of the practice distribution toward the student's competence boundary, where valid preference-pair yield is maximized. Experiments show that \method boosts yield across iterations and delivers stronger reasoning performance under an iso-effective training budget, demonstrating that it can distill more informative preference supervision for mathematical reasoning.
\end{abstract}

\section{Introduction}

\begin{figure*}[t]
\centering  
\makeatletter\let\@currentHref\@empty\makeatother
\includegraphics[width=0.97\textwidth]{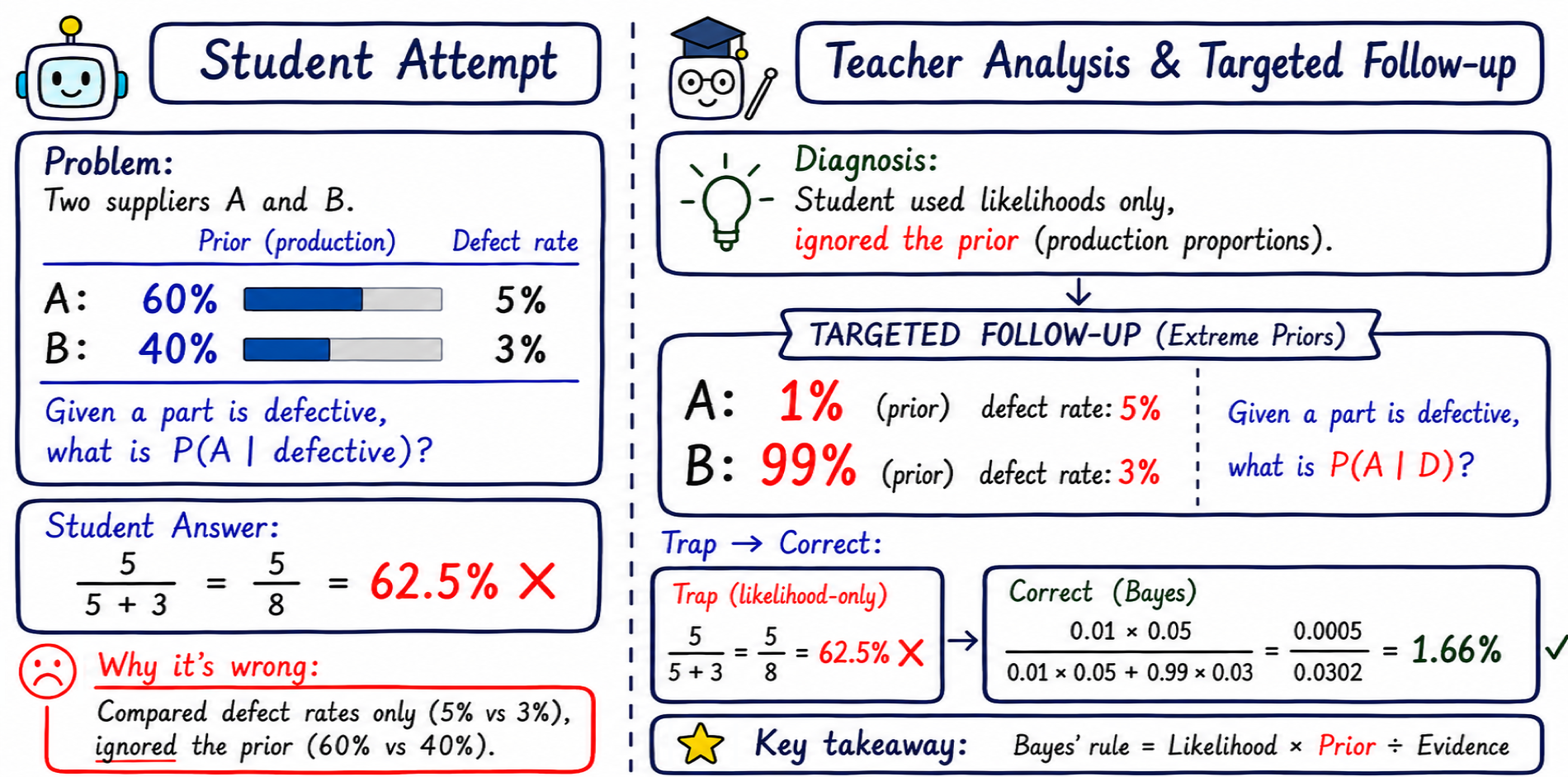}
\caption{\textbf{An illustrative example of \method.} 
\textit{Left:} The student directly compares likelihoods while ignoring prior probabilities, producing an incorrect posterior. 
\textit{Right:} \method diagnoses the misconception from the student's erroneous reasoning trace and generates a targeted follow-up with extreme priors that makes the error more salient, yielding more informative preference pairs for alignment.}
\label{fig:core_idea}
\end{figure*}

Artificial intelligence (AI) has advanced at an unprecedented pace in recent years, spanning natural language processing \citep{LLMSurvey}, computer vision \citep{zerovl,tian2026curvatureadaptiveconsistencyflowmatching}, multimodal understanding and generation \citep{geminiteam2025geminifamilyhighlycapable,yu2024samwav2lip}, and intelligent agents \citep{shinn2023reflexion,wang2026mapgraphprovenanceawaresharedmemory,wang2026agenttracestrustsurvey}. This progress has been driven by sustained advances in learning paradigms~\citep{ouyang2022training,wu2026toolaugmentedpolicyoptimizationsynergizing}, model architectures~\citep{vaswani2017attention,yu2026branch}, data scaling~\citep{yuan2023scalingrelationshiplearningmathematical}, and theoretical advances in understanding learning dynamics \citep{lai2023generalizationabilitywideresidual,yu2025divergence} and generalization \citep{li2024eigenvalue,chen2024OntheImpacts}. Within this broader development, Large Language Models (LLMs) have emerged as a central paradigm, achieving increasingly strong performance across a wide range of tasks and becoming a foundation for modern AI applications \citep{ouyang2022training}. Their rapid development has consequently attracted growing interest in how to improve their reasoning, instruction following, and interaction with users and environments.

Against this backdrop, the landscape of LLMs has undergone a dramatic shift toward heterogeneity, with model scales, architectures, and deployment regimes ranging from massive frontier systems (GPT: \citealp{openai2024gpt4technicalreport}; Llama: \citealp{llama3}) to specialized, lightweight variants (Phi-4-Mini: \citealp{microsoft2025phi4minitechnicalreportcompact}; Gemini Nano: \citealp{geminiteam2025geminifamilyhighlycapable}) optimized for edge deployment. This heterogeneity poses a challenge for model alignment: static training sets are increasingly unlikely to match the learning needs of models at different capability levels. Instances that are informative for one model may be too easy, too difficult, or even misleading for another, reducing the yield of supervision and potentially exacerbating hallucination-prone behavior \citep{gekhman-etal-2024-fine}.

Consequently, the alignment paradigm is shifting toward online, preference-based optimization loops~\cite{rafailov2023dpo, pang2024iterative}. By generating rollouts on-policy, these online loops ensure that the learning signal is calibrated to the current reasoning state of the student model. This capability is crucial for maintaining effective alignment across heterogeneous models. Under this paradigm, supervision typically arises from natural contrasts: the student samples multiple rollouts, and a preference pair is formed only when the set contains both correct and incorrect trajectories.
However, the efficacy of online rollouts remains fundamentally capped by the nature of the underlying prompt pool. Even if rollouts are generated on-policy and updated in real time, a fixed practice distribution may still fail to sample problems that match the model's current capability, thereby producing too few informative preference pairs and wasting computational resources.
Existing remedies often rely on brute-force scaling: increasing rollout width, expanding the static prompt pool, or using large-scale synthetic datasets~\cite{yu2023metamath,toshniwal2024openmathinstruct,chen2024controlmath}.

A seemingly more adaptive alternative is active learning, which selects high-uncertainty prompts from a candidate pool~\cite{settles2009active}. Yet for iterative preference optimization, active learning remains practice-agnostic in a crucial sense: it typically selects among existing prompts rather than generating targeted follow-ups tailored to the student model's current state. As a result, it can still produce redundant queries or abrupt difficulty shifts, worsening preference-pair scarcity rather than alleviating it.
This problem is conceptually a curriculum control problem. Classical curriculum learning and self-paced learning show that \emph{how} examples are scheduled can be as important as \emph{how many} examples are seen~\cite{bengio2009curriculum,kumar2010self}. However, applying curriculum ideas to iterative preference optimization introduces an additional challenge: the feedback signal is sparse, noisy at low sample counts, and non-stationary across iterations.

In this paper, we propose \textbf{\method}, a data-efficient framework for iterative preference distillation in mathematical reasoning. 
Our core insight is simple: \textit{conditioning teacher generation on student mistakes can improve both the yield of preference pairs and the quality of preference supervision.} As illustrated in Figure~\ref{fig:core_idea}, instead of treating the practice distribution as static, \method closes the loop between \emph{student outcomes} and \emph{next-iteration practice}.
It follows a \emph{Diagnose-and-Generate} paradigm: (i) diagnose how much usable preference signal the current curriculum yields; (ii) adaptively balance exploration and exploitation; (iii) allocate topic budgets using Empirical Bayes shrinkage to avoid low-count volatility~\cite{efron1973stein}; and (iv) generate targeted follow-up questions by conditioning a teacher on the student's recent mistakes, focusing computation near the decision boundary through uncertainty-weighted resampling rather than static pool selection~\cite{settles2009active}. 
This design increases the \emph{quantity} of effective preference pairs while improving the \emph{quality} of supervision by making new questions explicitly target observed failure modes.
Our contributions are as follows:     
\begin{itemize}[leftmargin=*,nosep,itemsep=1mm]
  \item We formalize iterative preference distillation in verifiable mathematical reasoning as a \textit{signal-yield} problem, and empirically show that static practice distributions can lead to a rapid decline in natural contrastive supervision.
  \item We present \textbf{\method}, which couples a signal-aware pacing rule with Empirical Bayes topic scoring to robustly steer generation toward high-yield regions without over-reacting to low-count noise.
  \item We demonstrate that \textit{mistake-conditioned generation} provides a simple and effective way to concentrate practice near the student's decision boundary, yielding higher-quality preference pairs with greater information density per unit budget than practice-agnostic generation.
\end{itemize}

\section{Related Work}

\paragraph{Iterative Preference Optimization in Reasoning}
Post-training for LLM alignment has expanded from online RLHF pipelines~\citep{schulman2017proximal,ouyang2022training} to simpler preference-based objectives such as DPO~\cite{rafailov2023dpo}, which are easier to scale into iterative loops. Recent iterative variants for reasoning (e.g., IRPO, \citealt{pang2024iterative}; SVPO, \citealt{chen2024step}) focus on improving the update rule or pair construction. Recent methods such as SAI-DPO dynamically select training data according to the model's evolving reasoning abilities~\citep{rao-etal-2026-dynamic}. In contrast, \method diagnoses preference-pair yield and further generates targeted practice from the student's erroneous reasoning traces. \method addresses an orthogonal bottleneck: even with a fixed verifier and rollout width, the practice distribution can quickly degenerate into all-correct or all-wrong regimes, starving the learner of usable preference pairs. In the spirit of curriculum learning and prioritized replay~\citep{bengio2009curriculum,kumar2010self,schaul2015prioritized,jiang2020plr}, our contribution is not a new objective but a signal-aware mechanism that allocates training budget where valid preference pairs are densest.

\paragraph{On-Policy Distillation (OPD)}
Recent on-policy distillation methods transfer teacher knowledge at the \emph{signal level}: OPSD~\citep{zhao2026self} and SDPO~\citep{hubotter2026reinforcement} let the same model play teacher and student under different contexts and minimise a token-level KL divergence, while OPCD~\citep{ye2026onpolicy} enriches the teacher representation with auxiliary context.
These approaches operate on fixed tasks or prompts and improve the supervisory distribution over the \emph{same} problem.
\method operates at an orthogonal, \emph{data level}: the teacher analyses the student's erroneous reasoning trace, diagnoses the underlying misconception, and synthesises a new question that forces the student to confront it, thereby reshaping the training distribution rather than the per-token supervision.

\paragraph{Reasoning Data Synthesis}
Throughout the lifecycle of large language models, data construction, curation, and feedback play central roles~\citep{rao2025data}.
Existing work has explored diverse data synthesis paradigms, including seed-based generation~\citep{wang2023self,xu2023wizardlm}, seedless generation~\citep{xu2025magpie}, and structured or design-guided synthesis~\citep{yu-etal-2026-mathagent,liu2026designer}. However, these approaches generally overlook the value of targeted feedback. In contrast, \method starts from the student's erroneous reasoning traces and uses targeted follow-up questions to continuously generate preference pairs with greater information content.

\paragraph{Mistake-Driven Augmentation}
Closest to our approach, LLM2LLM~\cite{lee2024llm2llm} prompts a teacher to generate new samples from student errors, but primarily conditions on the failed question rather than the student's erroneous reasoning trace. 
Meanwhile, \citet{rao2026miningsynthesisrethinkingexploration} improve exploration efficiency by mining and correcting failed reasoning trajectories, turning unsuccessful explorations into more informative training signals.
Beyond reasoning-data augmentation, recent agent research has emphasized the value of execution traces, failure histories, and provenance for diagnosing erroneous states and guiding corrective actions~\citep{yu2026faulty}. In contrast, \method does not repair persistent agent memory; instead, it uses erroneous reasoning traces to diagnose misconceptions and generate targeted follow-up questions, integrating this mistake-driven augmentation with iterative preference distillation. Rather than correcting existing failures, it creates new questions near the decision boundary to improve the efficiency and informativeness of contrastive supervision.

\section{Methods}
\label{sec:methods}
\begin{figure*}[t]
    \centering
    \includegraphics[width=0.99\linewidth]{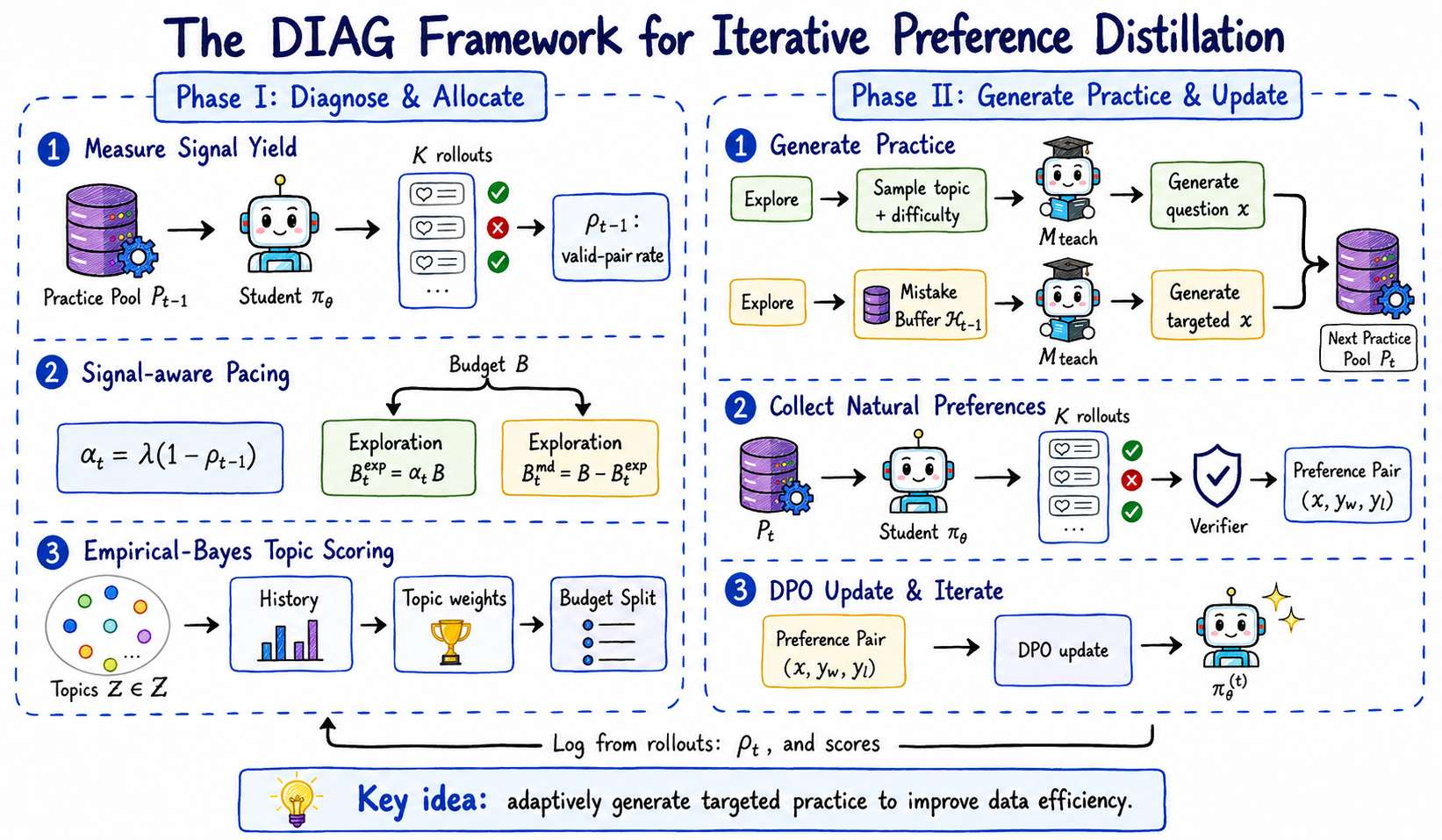}
    \caption{\textbf{The \method Framework} for Iterative Preference Distillation. \method identifies the student's competence boundary and uncertainty, and generates targeted practice to sustain informative preference pairs, improving both supervision quality and data efficiency for math reasoning alignment.}
    \label{fig:main}
\end{figure*}

\subsection{Overview}
As shown in Figure~\ref{fig:main}, we present \textbf{\method}, a framework that improves data efficiency in iterative preference distillation for mathematical reasoning. Rather than training on a fixed prompt pool (\emph{Static Practice}), \method adopts a \emph{Diagnose-and-Generate} paradigm: after each DPO iteration it uses the student's failures to adaptively reshape the practice distribution for the next round. 

Concretely, the method operates in two coupled phases (see Algorithm~\ref{alg:ace_dpo} for the overall procedure):
\begin{itemize}[nosep,itemsep=1mm]
\item[(1)] \textit{Diagnose \& Allocate}: estimate the effective pair yield from the previous iteration, set an exploration--exploitation budget split, and assign topic-level quotas via utility estimates smoothed by Empirical Bayes shrinkage to avoid noisy swings in the low-sample regime.
\item[(2)] \textit{Generate Practice}: perform weighted resampling over recent mistakes within each topic and query a teacher to generate new follow-up problems conditioned on the failed question and the student's incorrect response.
\end{itemize}

\begin{algorithm}[tb]
\caption{\method Distillation Loop}
\label{alg:ace_dpo}
\begin{algorithmic}[1]
\STATE {\bfseries Input:} Topic set $\mathcal{Z}$, Difficulty levels $\mathcal{D}$, teacher $\mathcal{M}_{\text{teach}}$, student $\pi_\theta$, rollout width $K$, per-iter budget $B$
\STATE {\bfseries Init:} $N_z\!\leftarrow\!0,\; C_z\!\leftarrow\!0\;\forall z$;\; $\mathcal{H}_0\!\leftarrow\!\emptyset$;\; $\alpha_1\!\leftarrow\!1$
\FOR{iteration $t=1$ {\bfseries to} $T$}
    \STATE \textit{\color{gray}// Phase I: Diagnose \& Allocate}
    \STATE $B^{\text{rnd}}_t \leftarrow \lfloor \alpha_t B \rfloor$,\quad $B^{\text{exp}}_t \leftarrow B - B^{\text{rnd}}_t$
    \STATE Compute scores $v_z$;\quad $\pi_{\text{sampler}}(z) \propto v_z$
    \STATE $\mathcal{D}_t \leftarrow \emptyset$,\; $P_t \leftarrow \emptyset$, \; $\mathcal{H}_t \leftarrow \emptyset$
    \STATE \textit{\color{gray}// Phase II: Generate Practice}
    \FOR{$j=1$ {\bfseries to} $B$}
        \IF{$j \le B^{\text{rnd}}_t$}
            \STATE Sample $(z,d)\sim \mathcal{U}(\mathcal{Z}\times\mathcal{D})$ \hfill \textit{\color{gray}$\triangleright$ explore}
            \STATE $x \sim \mathcal{M}_{\text{teach}}(\cdot \mid z,\, d)$
        \ELSE
            \STATE Sample $z\sim \pi_{\text{sampler}}$;\;
            \STATE Get $(x_0,d,y_{\text{err}})\!\sim\!\mathcal{H}_{t-1}|_z$ \hfill \textit{\color{gray}$\triangleright$ exploit}
            \STATE $x \sim \mathcal{M}_{\text{teach}}(\cdot \mid z,\, d,\, x_0,\, y_{\text{err}})$
        \ENDIF
        \STATE $a^\star \leftarrow \mathcal{M}^{\text{ans}}_{\text{teach}}(x)$ 
        \hfill \textit{\color{gray}$\triangleright$ teacher solves}
        \STATE Roll out $\{y_k\}_{k=1}^K \sim \pi_\theta(\cdot\mid x)$;\; $r_k \leftarrow \mathcal{V}(y_k; x,a^\star)$
        \STATE $P_t \leftarrow P_t \cup \{x\}$;\; $N_z \leftarrow N_z + 1$
        \IF{$\exists\, i,j:\; r_i\!=\!1,\; r_j\!=\!0$}
         \STATE $y_w \leftarrow y_i,\quad y_l \leftarrow y_j$ \
            \STATE $\mathcal{D}_t \leftarrow \mathcal{D}_t \cup \{(x,y_w,y_l)\}$
            \STATE $C_z \leftarrow C_z + 1$
            \STATE $\mathcal{H}_t \leftarrow \mathcal{H}_t \cup \{(x,z,d,y_{{l}})\}$
        \ENDIF
    \ENDFOR
    \STATE Update $\pi_\theta$ by minimizing $\mathcal{L}_{\text{DPO}}$ on $\mathcal{D}_t$
    \STATE $\rho_t \leftarrow |\{x\!\in\!P_t : \mathbb{I}_{\text{valid}}(x)\!=\!1\}|\,/\,|P_t|$
    \STATE $\alpha_{t+1} \leftarrow \lambda(1-\rho_t)$ \hfill \textit{\color{gray}$\triangleright$ next-iter explore rate}
\ENDFOR
\end{algorithmic}
\end{algorithm}

\subsection{Preliminaries: Natural Preference Signal}
Let $\pi_\theta$ denote the student policy, and let
$V(y;x)\in\{0,1\}$ denote a task verifier.
For each prompt $x$, we draw $K$ stochastic rollouts $$\{y^{(k)}\}_{k=1}^K \sim \pi_\theta(\cdot \mid x)$$ and obtain correctness labels $\{r^{(k)}\}_{k=1}^K$, where $r^{(k)}=V(y^{(k)};x)$.

\paragraph{Natural Pair Existence}
A prompt $x$ is \emph{valid} for constructing preference pairs if its rollouts contain at least one correct and one incorrect response:
\begin{align}
\label{eq:valid_pair}
\mathbb{I}_{\text{valid}}(x) \triangleq \mathbb{I}\Big(\exists i,j:\ r^{(i)}=1\ \land\ r^{(j)}=0\Big).
\end{align}
Define the signal yield of a practice pool $P_t$ as
\begin{equation}
\label{eq:yield}
\rho_t \triangleq
\frac{1}{|P_t|}
\sum_{x\in P_t}
\mathbb{I}_{\text{valid}}(x).
\end{equation}
When $\rho_t$ collapses (e.g., many prompts yield all-correct or all-wrong rollout sets), iterative preference learning experiences gradient starvation. \method explicitly adapts the practice generation process to maintain high $\rho_t$ over iterations.

\subsection{Phase I: Diagnose \& Allocate}
\label{sec:phase1}
At each iteration, \method first diagnoses the amount of \emph{usable preference signal} produced by the current practice distribution, then allocates the generation budget across topics accordingly. This phase combines signal-aware pacing with Empirical Bayes topic scoring to produce an adaptive, targeted, and noise-resilient generation plan.

\paragraph{Signal-aware pacing}
At the beginning of iteration $t$, \method uses the yield measured on the
previous practice pool $P_{t-1}$ to set the exploration rate:
\begin{equation}
\label{eq:pacing_alpha}    
\alpha_t \triangleq \lambda\big(1-\rho_{t-1}\big),
\end{equation}
so that exploration increases when informative natural pairs are scarce and decreases when the current curriculum already produces abundant on-policy contrastive signal. 
By construction, $\alpha_t\in[0,\lambda]\subseteq[0,1]$, where $\lambda\in(0,1]$ is a fixed scaling factor controlling how aggressively the sampler shifts budget in response to the measured yield.

\paragraph{Budget split}
Given a fixed generation budget of $B$ questions per iteration, we allocate
\begin{equation}
\label{eq:budget_split}
B^{\text{rnd}}_{t} \triangleq \left\lfloor \alpha_{t} B \right\rfloor,\quad
B^{\text{exp}}_{t} \triangleq B - B^{\text{rnd}}_{t},
\end{equation}
where $B^{\text{rnd}}_{t}$ is assigned to \emph{uniform exploration} over the topic--difficulty space $(z,d)$, and $B^{\text{exp}}_{t}$ is assigned to the \emph{exploitation} stream. 
Intuitively, when $\rho_{t-1}$ is low, we expand coverage to search for regions that can generate usable preference signal; when $\rho_{t-1}$ is high, we concentrate the exploitation budget on high-yield topics and recent mistakes to maximize sample efficiency. In practice, this acts as a damping mechanism: when the model encounters a ``plateau'' with low signal yield, the system automatically allocates more budget to broad exploration to discover new learnable frontiers. See Appendix~\ref{app:implementation} for details.

\paragraph{Empirical Bayes topic scoring}
We maintain per-topic statistics to estimate which topics are most likely to yield informative natural preference pairs, and use these estimates to guide the allocation of the exploitation budget $B^{\text{exp}}_{t}$. 
Let $z \in \mathcal{Z}$ denote a topic and let $x$ be a question tagged with topic $z$. 
For each attempted question, we update 
\begin{equation}
\label{eq:NzCz_def}
N_z \leftarrow N_z + 1,\qquad 
C_z \leftarrow C_z + \mathbb{I}_{\text{valid}}(x),
\end{equation}
where $N_z$ counts how many questions of topic $z$ have been attempted, while $C_z$ counts how often topic $z$ produced a usable natural contrastive signal.

We model the latent utility $p_z$ (the probability that topic $z$ yields a valid natural pair) as a Beta random variable with Empirical Bayes prior: 
\begin{equation*}
\begin{aligned}
p_z &\sim \mathrm{Beta}(\alpha_0,\beta_0), \\
p_z | (C_z,N_z) &\sim \mathrm{Beta}(\alpha_0+C_z, \beta_0+N_z-C_z).
\end{aligned}
\end{equation*}
Here, $\alpha_0$ and $\beta_0$ are prior pseudo-counts controlling the amount of shrinkage toward the global valid rate. Let $\bar{p}$ denote the global mean valid rate across topics, and let $\kappa>0$ denote the total prior strength. We set $\alpha_0=\kappa \bar{p}$ and $\beta_0=\kappa (1-\bar{p})$. 
The posterior mean gives a Stein-type shrinkage score:
\begin{equation}
\label{eq:eb_score}
v_z \triangleq \mathbb{E}[p_z\mid C_z,N_z] = \tfrac{C_z + \kappa \bar{p}}{N_z + \kappa }.
\end{equation}

\paragraph{Topic sampling policy}
We sample topics for the exploitation stream proportionally to $v_z$:
\begin{equation}
\label{eq:topic_sampler}
\pi_{\text{sampler}}(z) = {v_z}/{\textstyle{\sum_{z'\in\mathcal{Z}}} v_{z'}}.
\end{equation}
This Empirical Bayes scheduler prevents the curriculum from over-reacting to low-count noise while focusing generation on topics that repeatedly yield informative on-policy supervision.

\subsection{Phase II: Generate Practice}
\label{sec:phase2}

Given the budget split and topic sampler from Phase~I, we next construct the practice pool for the following iteration. The exploration stream samples uniformly, while the exploitation stream conditions on the student's recent failures to generate targeted follow-up questions.

\paragraph{Mistake buffer}
At iteration $t$, we collect a buffer of boundary mistake cases:
\begin{equation}
\mathcal{H}_t \triangleq \{(x,\,z,\,d,\,y_{\mathrm{err}})\},
\end{equation}
where $(z,d)$ denotes topic and difficulty, and $y_{\mathrm{err}}$ is an incorrect trace from a prompt that also admitted a correct rollout (i.e., a valid preference pair). Only such mixed-outcome prompts enter $\mathcal{H}_t$, as they lie near the student's competence boundary. Degenerate episodes still update the signal-yield and topic statistics; when their mass grows, the pacing rule shifts budget toward exploration.

\paragraph{Trace-aware question synthesis}
For each exploitation slot, we first sample a topic $z \sim \pi_{\text{sampler}}(z)$, then retrieve a recent mistake case $(x,z,d,y_{\text{err}})\in\mathcal{H}_{t-1}$ with the same topic (and matched difficulty), via uncertainty-weighted resampling \citep{settles2009active},
and query the teacher to generate a \emph{targeted variant}:
\begin{equation}    
\label{eq:teacher_targeted}
x' \sim \mathcal{M}_{\text{teach}}(\cdot \mid z,\ d,\ x,\ y_{\text{err}},\ \text{instruction}).
\end{equation}
The instruction enforces: (i) preserve the underlying concept (same $z$), (ii) match difficulty $d$, and (iii) alter surface form (numbers/context) to avoid memorization while probing the same failure mode.
For the exploration stream, questions are generated from metadata alone without mistake conditioning. The resulting questions together form the next practice pool $P_{t}$.

\paragraph{Theoretical interpretation}
Intuitively, generating from mistakes focuses computation near the student's competence boundary. Appendix~\ref{app:yield-geometry} formalizes this: for rollout width $K$, the probability that a prompt yields a valid preference pair is $1-p^K-(1-p)^K$, which peaks at $p=1/2$ and vanishes in the all-correct and all-wrong regimes. Appendix~\ref{app:kl-reweighting} then derives a KL-regularized reweighting that exponentially upweights high-yield prompts. \method approximates this ideal update via Empirical-Bayes topic allocation and teacher generation conditioned on recent boundary mistakes.

\paragraph{Policy update}
Once the practice pool $P_{t}$ is constructed via Phases I--II, we collect preference pairs by labeling student rollouts based on ground-truth correctness. The student policy $\pi_\theta$ is then updated by minimizing the standard DPO objective:
\begin{equation}
\label{eq:dpo_obj}
\begin{split}
&\mathcal{L}_{\text{DPO}}(\pi_\theta, \pi_{\text{ref}}) = -\mathbb{E}_{(x,y_w,y_l)\sim\mathcal{D}}  \\
&\qquad\Big[\log \sigma\big(\beta[h_\theta(x,y_w) - h_\theta(x,y_l)]\big)\Big],  
\end{split}
\end{equation}
where $h_\theta$ is the implicit reward difference and $\pi_{\text{ref}}$ is a fixed reference policy for the current iteration.

\begin{table*}[t]
\centering
\resizebox{0.99\textwidth}{!}{     
\begin{tabular}{lccccccccc}
\toprule
\multirow{2}{*}{\textbf{\centering Method}} & \multicolumn{3}{c}{\textbf{Standard}} & \multicolumn{2}{c}{\textbf{Exam}} & \multicolumn{3}{c}{\textbf{Competition (Hard)}} & \multirow{2}{*}{\textbf{\centering Avg.}}\\
\cmidrule(lr){2-4} \cmidrule(lr){5-6} \cmidrule(lr){7-9}
& \parbox{1.3cm}{\centering\small\textbf{GSM8K}}  
& \parbox{1.3cm}{\centering\small{\textbf{MATH}}\\\centering\small{\textbf{500}}} 
& \parbox{1.3cm}{\centering\small{\textbf{Minerva}}\\\centering\small{\textbf{Math}}} 
& \parbox{1.3cm}{\centering\small\textbf{Gaokao}\\\centering\small{\textbf{2023en}}}  
& \parbox{1.3cm}{\centering\small\textbf{College}\\\centering\small\textbf{Math}} 
& \parbox{1.3cm}{\centering\small\textbf{Olympiad}\\\centering\small\textbf{Bench}}  
& \parbox{1.3cm}{\centering\small\textbf{AIME24}\\\centering\small\textbf{(Avg@8)}} 
& \parbox{1.3cm}{\centering\small\textbf{AMC23}\\\centering\small\textbf{(Avg@8)}}
& \\
\midrule
\multicolumn{10}{c}{\textbf{\textit{Qwen2.5-Math-7B}}} \\
\midrule
\textsc{Base} (Start) & 84.2 & 70.2 & 26.5 & 58.7 & 35.0 & 33.8 & {24.2} & 54.7 & 48.4 \\
\textsc{Numina} $\rightarrow$ IDPO & 85.4 & 70.4 & 29.8 & 59.0 & 37.8 & 35.9 & 22.5 & 56.6 & 49.7 \\
\textsc{Static Gen} $\rightarrow$ IDPO & 86.5 & 69.6 & 33.8 & 60.8 & 38.7 & 35.7 & 22.1 & 59.7 & 50.9 \\
\rowcolor{gray!10} \textbf{\textsc{\method} (Ours)} & \textbf{87.1} & \textbf{70.8} & \textbf{34.9} & \textbf{61.0} & \textbf{41.9} & \textbf{38.4} & \textbf{25.0} & \textbf{63.1} & \textbf{52.8} \\
\midrule
\multicolumn{10}{c}{\textbf{\textit{Qwen3-8B-Base}}} \\
\midrule
\textsc{Base} (Start) & 89.1 & 70.0 & 35.3 & 57.4 & 36.3 & 33.9 & 8.8 & 50.0 & 47.6 \\
\textsc{Numina} $\rightarrow$ IDPO & 90.9 & 73.4 & 34.9 & 62.3 & 38.1 & 40.1 & 8.3 & 56.2 & 50.5 \\
\textsc{Static Gen}  $\rightarrow$ IDPO & 91.1 & 76.6 & \textbf{42.6} & 63.1 & {39.1} & 40.0 & 10.4 & 52.2 & 51.9 \\
\rowcolor{gray!10} \textbf{\textsc{\method} (Ours)} & \textbf{92.0} & \textbf{79.0} & {40.8} & \textbf{64.7} & \textbf{39.7} & \textbf{41.3} & \textbf{14.6} & \textbf{57.5} & \textbf{53.7} \\
\midrule
\multicolumn{10}{c}{\textbf{\textit{Llama-3.1-8B-Instruct}}} \\
\midrule
\textsc{Base} (Start) & 84.8 & 47.6 & 25.7 & 41.6 & 28.8 & 14.5 & 7.1 & 21.2 & 33.9 \\
\textsc{Numina} $\rightarrow$ IDPO & 86.3 & \textbf{50.8} & 27.9 & 43.4 & 28.9 & 15.7 & 5.8 & 24.4 & 35.4 \\
\textsc{Static Gen}  $\rightarrow$ IDPO & 86.9 & {50.2} & 29.4 & 42.1 & 30.0 & 15.6 & 6.7 & 26.6 & 35.9 \\
\rowcolor{gray!10} \textbf{\textsc{\method} (Ours)} & \textbf{88.4} & {48.2} & \textbf{31.6} & \textbf{45.2} & \textbf{32.1} & \textbf{17.6} & \textbf{7.5} & \textbf{28.4} & \textbf{37.4} \\
\bottomrule
\end{tabular}
}
\caption{\textbf{Main Results on Mathematical Reasoning (Fixed Total Budget).}\label{tab:main_results}
We compare four data sources under the \emph{same total student-sample budget}:
(\textsc{Base}) the starting checkpoint,
(\textsc{Numina}) IDPO on a static open-source pool,
(\textsc{Static Gen}) IDPO on a fixed teacher-generated pool conditioned on (topic, difficulty),
and \textbf{\method (Ours)} IDPO on online teacher generation conditioned on (failed question, wrong answer, topic, difficulty).}
\label{tab:main_results_ace_fullsuite_grouped}
\end{table*}

\section{Experiments}
\label{sec:experiments}
\subsection{Setup}

\paragraph{Benchmarks}
We evaluate on two standard math benchmarks
GSM8K~\cite{gsm} and MATH500~\cite{hendrycks2021measuringmathematicalproblemsolving}, together with harder testbeds:
MinervaMath~\cite{lewkowycz2022solvingquantitativereasoningproblems}, Gaokao2023En~\cite{liao-etal-2024-mario}, College Math~\cite{tang2024mathscale}, 
OlympiadBench~\cite{he-etal-2024-olympiadbench}, AIME24~\citep{aime}, and AMC23~\citep{amc}.
We use greedy decoding for all benchmarks except AMC23 \& AIME24, whose small test sizes make single-run accuracy more sensitive to sampling noise. For these two benchmarks, we sample 8 solutions per problem with temperature $T=0.1$ and top-$p=0.95$, and report the mean pass@1 accuracy to reduce evaluation variance.

\paragraph{Student Models}
We evaluate three open-weight student models with approximately 7B--8B parameters: Qwen2.5-Math-7B~\cite{yang2024qwen25mathtechnicalreportmathematical}, Qwen3-8B-Base~\cite{yang2025qwen3}, and Llama-3.1-8B-Instruct~\cite{llama3}, spanning multiple model series and capability profiles.

\paragraph{Budget and fairness}
All methods share the same total student-sample budget. 
We run IDPO~\citep{guo2024direct,pang2024iterative} for $7$ iterations with $3\times 10^3$ prompts per iteration, keeping the decoding protocol identical across methods. We employ Qwen3-235B (Int4) as the teacher model.

\paragraph{Practice pool construction (what differs)}
We compare three pool-construction strategies under the same student-sample budget:
\begin{itemize}[leftmargin=*,nosep]
    \item \textbf{Static Dataset Pool (Numina).} We use the open-source NuminaMath dataset~\cite{li2024numinamath} as a fixed practice pool, without online regeneration.
    \item \textbf{Static Gen (Teacher).} The teacher generates all $T \times B$ questions upfront using only metadata such as topic and difficulty. The resulting pool is fixed and partitioned equally across iterations. 
    \item \textbf{Ours: Mistake-Conditioned.} At each iteration, the teacher regenerates the pool by conditioning on the student's failures (Sec.~\ref{sec:methods}).     
\end{itemize}

\subsection{Main Results}

Table~\ref{tab:main_results} reports the full results.
Despite identical student-sample budgets, \method achieves the strongest average performance across all three student models.
The gains are most pronounced on harder testbeds, where generic practice pools are more likely to enter degenerate regimes, reducing the availability of natural preference supervision.

\begin{table}[t]
\centering
\resizebox{0.46\textwidth}{!}{
\begin{tabular}{lccc}
\toprule
\multirow{2}{*}{\textbf{Task}} &
\multicolumn{3}{c}{\textbf{Sampling Strategies}} \\
\cmidrule(lr){2-4}
& \textbf{w/o Explore} & \textbf{w/o EB} & \textbf{Ours} \\
\midrule
\rowcolor{gray!8} {GSM8K}        & \textbf{87.6} & 86.4 & 87.1 \\
{MATH500}     & 70.4 & 70.6 & \textbf{70.8} \\
\rowcolor{gray!8} {MinervaMath}   & \textbf{35.7} & 35.3 & 34.9 \\
{Gaokao2023}   & 59.5 & 59.7 & \textbf{61.0} \\
\rowcolor{gray!8} {Olympiad}   & 37.5 & \textbf{38.7} & 38.4 \\
{CollegeMath}  & \textbf{42.4} & 40.3 & 41.9 \\
\rowcolor{gray!8} {AIME24}     & 24.2 & 24.2 & \textbf{25.0} \\
{AMC23}      & 58.1 & 59.4 & \textbf{63.1} \\
\midrule
\rowcolor{gray!20} \textbf{Avg.} & 51.9 (-0.9) & 51.8 (-1.0) & \textbf{52.8} \\
\bottomrule
\end{tabular}}
\caption{\textbf{Ablation on Sampling\,/\,Scheduling.}
We ablate the uniform exploration stream (\textbf{w/o Explore}) and EB smoothing (\textbf{w/o EB}) under a fixed rollout budget.}
\label{tab:ablation_sampling}
\end{table}
\subsection{Ablation Study}
\paragraph{Sampling \& Scheduling}
We ablate the sampling \& scheduling design on Qwen2.5-Math-7B under a fixed rollout budget: 
(i) \textbf{w/o Explore}: remove the uniform exploration stream, allocating the entire budget to score-based exploitation;
(ii) \textbf{w/o Empirical Bayes (EB)}: replace EB shrinkage with raw per-topic rates;
(iii) \textbf{Ours}: the full scheduler.

Table~\ref{tab:ablation_sampling} shows that both components matter in different ways. Removing exploration yields comparable scores on standard benchmarks but exhibits brittleness on competition-style evaluation (AMC23: 58.1 vs 63.1). Removing EB shrinkage leads to broad degradation on most benchmarks (Avg 51.8), confirming that naive frequency-based scoring is unstable in the low-sample regime.

\begin{table}[t]
\centering
\resizebox{0.48\textwidth}{!}{
\begin{tabular}{lcc}
\toprule
\textbf{Method / Variant} & \textbf{Avg.} & \textbf{$\Delta$} \\
\midrule
\rowcolor{gray!20} \textbf{Full \method} & \textbf{52.8} & -- \\
\midrule
\rowcolor{gray!8} \textit{Impact of conditioning on failed question} &  & \\
\quad w/o Wrong-QA (no failed context) & 51.3 & {-1.5} \\
\midrule
\rowcolor{gray!8} \multicolumn{2}{l}{\textit{Impact of conditioning on wrong answer/trace}} & \\
\quad w/o Wrong-A (no wrong answer) & 52.1 & {-0.7} \\
\bottomrule
\end{tabular}
}
\caption{\textbf{Ablation on Teacher Conditioning Signals.} 
$\Delta$ denotes performance drop relative to the full method.}
\label{tab:ablation_mistake}
\end{table}

\paragraph{Teacher Conditioning Signals}
We isolate which components of the teacher conditioning signal matter for targeted generation, all on Qwen2.5-Math-7B with identical budget and topic/difficulty controls:
(i) \textbf{w/o Wrong-QA}: teacher conditioned only on topic/difficulty (no failed context);
(ii) \textbf{w/o Wrong-A}: teacher receives the failed question but not the student's wrong answer;
(iii) \textbf{Ours}: full conditioning on topic + difficulty + failed question + wrong answer.

As shown in Table~\ref{tab:ablation_mistake}, performance increases monotonically as richer error context is provided.
Removing the failed context ($-1.5$) weakens the teacher's ability to anchor generation to the exact skill being tested.
Removing the wrong answer ($-0.7$) confirms that error-aware generation is not equivalent to simple topic-conditioned rewriting: observing the student's mistake helps the teacher probe the same misconception more reliably.

\subsection{Cross-Task Generalization}
\label{sec:cross_task}
To verify that \method does not incur an alignment tax, we evaluate Qwen3-8B-Base on BBH~\citep{suzgun2023challenging}, HumanEval~\citep{chen2021evaluatinglargelanguagemodels}, MMLU~\citep{hendrycks2021measuring}, and TruthfulQA~\citep{lin-etal-2022-truthfulqa}. 
Table~\ref{tab:cross_task_generalization} shows that \method maintains comparable performance on knowledge and truthfulness benchmarks while showing mild positive transfer on reasoning and code generation. 

\begin{table}[h!]
\centering
\setlength{\tabcolsep}{1.5mm}
\resizebox{0.48\textwidth}{!}{
\begin{tabular}{lccccc}
\toprule
\multirow{2}{*}{\textbf{\centering Method}} & \multirow{2}{*}{\textbf{BBH}} &
\multirow{2}{*}{\parbox{1.2cm}{\centering\textbf{Human}\\[-1mm]\centering\textbf{Eval}}} &
\multirow{2}{*}{\textbf{MMLU}} & \multicolumn{2}{c}{\textbf{TruthfulQA}} \\
\cmidrule(lr){5-6}
 & & & & \small\textbf{MC1} & \small\textbf{MC2} \\
\midrule
\rowcolor{gray!8} Qwen3-8B-Base  & 75.5 & 64.6 & 76.9 & 33.5 & 51.1 \\
\quad+~Static Gen     & ~ 75.6$\,^\uparrow$ & 64.6 & 76.9 & ~\,\,33.6$\,^\uparrow$ & ~\,\,\textbf{51.2}$\,^\uparrow$ \\
\midrule
\rowcolor{gray!20} \quad+~\method      & ~\,\textbf{75.6}$\,^\uparrow$ & ~\,\,\textbf{64.7}$\,^\uparrow$ & \textbf{76.9} & ~\,\,\textbf{33.9}$\,^\uparrow$ & 51.0 \\
\bottomrule
\end{tabular}
}
\caption{\textbf{Cross-Task Generalization and Alignment Tax Evaluation.}
\textbf{Bold} denotes the best result; $^\uparrow$ indicates improvement over the base model.}
\label{tab:cross_task_generalization}
\end{table}

\begin{figure}[t!]
    \centering
    \includegraphics[width=0.87\linewidth]{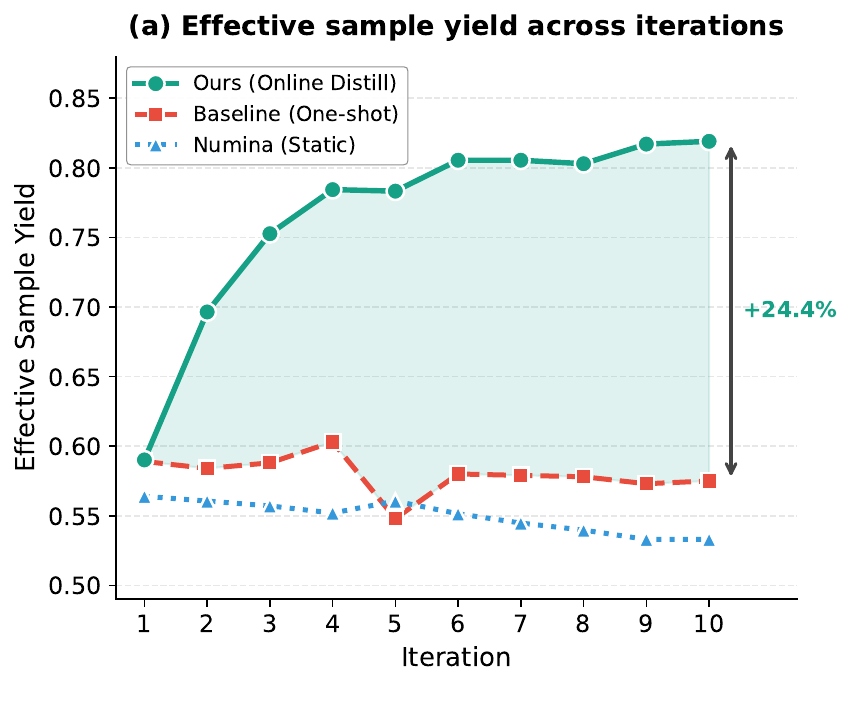}
    \vspace{-4mm}
    \caption{\textbf{Effective sample yield across iterations.} \method maintains a higher signal yield $\rho_t$ than baselines and avoids degeneration into uninformative regimes.}
    \label{fig:yield_dynamics}
\end{figure}

\section{Analysis}
\label{sec:analysis}

\subsection{Effective Sample Yield over Training}
\label{subsec:analysis_yield}

A central claim of our method is that optimizing the practice distribution improves sample efficiency under a fixed rollout budget. Following \eqref{eq:yield}, we track the effective yield $\rho_t$, defined as the fraction of questions whose $K$ rollouts contain both correct and incorrect outcomes, across iterations on Qwen2.5-Math-7B.

Figure~\ref{fig:yield_dynamics} reports an extended 10-iteration diagnostic run: our method increases $\rho_t$ from $\sim$0.59 to $\sim$0.83 (+24 pp) over 10 iterations, whereas both Static Gen and static Numina gradually decline to around 0.55, confirming that targeted generation keeps the training distribution near the student's moving competence boundary.

\subsection{Pass-Rate Distribution}
\label{subsec:analysis_passrate}

\begin{figure}[t]
    \centering
    \includegraphics[width=0.8\linewidth]{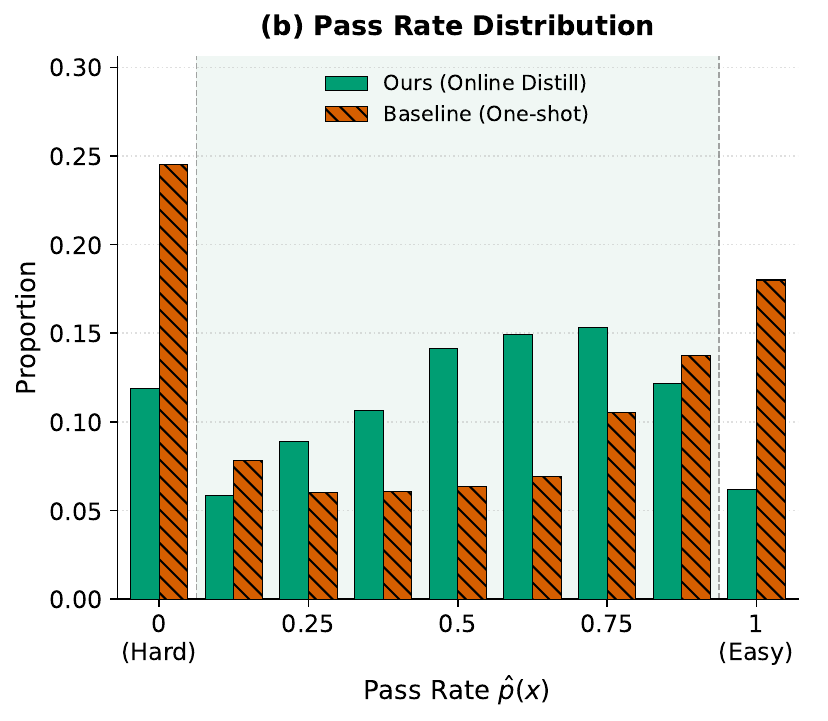}
    \vspace{-1mm}
    \caption{\textbf{Pass-rate distribution.} Our method concentrates more mass in the intermediate region, reducing both all-correct and all-wrong degeneracy.}
    \label{fig:passrate_dist}
\end{figure}

The yield gap above suggests that the key advantage comes from \emph{where} the model practices. We analyze the pass-rate distribution by estimating $\hat{p}(x) \triangleq \frac{1}{K}\sum_{k=1}^K r_k(x)$ for each generated question. Mass near $\hat{p}(x)\approx 1$ or $\hat{p}(x)\approx 0$ indicates wasted budget on trivial or intractable questions, respectively; both extremes cause gradient starvation. 
Figure~\ref{fig:passrate_dist} shows that our method concentrates more mass in the intermediate band, where the student is neither always correct nor always wrong. This suggests that mistake-conditioned generation acts as a calibration operator: it improves \emph{difficulty matching} rather than simply making questions harder.

\subsection{Iso-Effective Comparison}
\label{subsec:analysis_isoeff}

Our sampler produces more mixed episodes, which could confound improvements due purely to data quantity. To disentangle quantity from quality, we subsample each method to the same number ($\sim$8k) of effective training instances and train the same Qwen2.5-Math-7B student.
We additionally include two error-driven baselines under the same DPO framework:
{LLM2LLM-style}~\citep{lee2024llm2llm}, which generates new questions from student failures but without yield-aware scheduling (originally SFT; we use DPO for fair comparison);
and {SPIN}~\citep{chen2024self}, which constructs self-play preference pairs on a fixed prompt pool.

As shown in Table~\ref{tab:isoeffective}, our method remains consistently better even when the effective sample count is held constant.
SPIN (+0.5), operating on fixed prompts, confirms that prompt-level curriculum matters beyond response-level contrast.
LLM2LLM-style (+1.7) benefits from error-driven generation but conditions the teacher only on the failed question; \method's additional conditioning on the student's wrong reasoning trace enables follow-ups that target the specific misconception, and its EB-guided scheduling further improves signal yield.
This gap is consistent with the analysis in Appendix~\ref{app:yield-geometry} and~\ref{app:dpo-curvature}: boundary prompts produce more mixed rollouts, placing the model in a higher-curvature, more informative update region.

\begin{table}[t]    
\centering
\resizebox{0.48\textwidth}{!}{%
\renewcommand{\arraystretch}{1.15}
\begin{tabular}{lccc}
\toprule
\textbf{Method} & \textbf{Eff. Samples} & \textbf{Avg.} & \textbf{$\Delta$} \\
\midrule
Qwen2.5-Math-7B & -- & 48.4 & -- \\
\midrule
~~+ SPIN \cite{chen2024self} & 8k & 48.9 & +0.5 \\
\rowcolor{gray!10} ~~+ NuminaMath \cite{li2024numinamath} & 8k & 49.0 & +0.6 \\
~~+ Static IDPO \cite{zhangonline}      & 8k & 49.7 & +1.3 \\
\rowcolor{gray!10} ~~+ LLM2LLM \cite{lee2024llm2llm} & 8k & 50.1 & +1.7 \\
~~\textbf{+ Ours} & 8k & \textbf{50.7} & \textbf{+2.3} \\
\bottomrule
\end{tabular}
}
\caption{\textbf{Iso-effective comparison.} All methods use $\sim$8k effective instances and the same gradient steps. LLM2LLM-style adopts error-driven generation but trains with DPO for fair comparison.}
\label{tab:isoeffective}
\end{table}
\section{Conclusion}
We showed that the dominant bottleneck in iterative preference distillation for math reasoning is signal scarcity: as the student improves, static practice pools degenerate into uninformative regimes. To address this, we proposed \method, a Diagnose-and-Generate framework that reshapes the practice distribution via signal-aware pacing, Empirical-Bayes topic prioritization, and mistake-conditioned generation. Across diverse students and challenging math evaluations, \method consistently improves both data efficiency and final reasoning performance, with gains persisting even under iso-effective training, indicating higher-quality supervision beyond increased signal yield.
\section*{Limitations}
\method diagnoses signal scarcity in real time via Empirical Bayes shrinkage and uses mistake-conditioned generation to produce targeted follow-up problems near the student model's moving competence boundary.
Two main limitations remain:
(i)~the Diagnose-and-Generate loop relies on a capable teacher model for question synthesis, which may not be available or affordable in all settings;
(ii)~our evaluation is restricted to mathematical reasoning with reliable verifiers, and it is unclear how well the framework transfers to domains where correctness cannot be automatically checked.

\section*{Ethics Statement}
This work complies with the ACL Ethics Policy.
All experiments are conducted using publicly available datasets and widely accessible models, with proper citations provided throughout. Our method does not involve human subjects, private data, or any form of data annotation beyond automated answer verification. We acknowledge that LLMs, including those trained with our framework, may still produce incorrect or misleading reasoning steps. While DIAG improves mathematical reasoning reliability, users should independently verify model outputs before relying on them in high-stakes or real-world decision-making scenarios.

\bibliography{example_paper}

\newpage
\appendix

\section{Theoretical Perspective: Signal Yield and Practice-Space Reweighting}
\label{app:theory}

This appendix provides a compact theoretical view of DIAG. 
The main idea is that useful natural preference pairs are most likely to arise near the student's competence boundary, where the model sometimes succeeds and sometimes fails. 
DIAG can then be interpreted as a teacher-mediated approximation to an ideal practice-distribution reweighting that moves probability mass toward such high-yield regions.

\subsection{Signal Yield Peaks at the Competence Boundary}
\label{app:yield-geometry}

Let $\pi_\theta(y\mid x)$ be the student policy and let $V(y;x)\in\{0,1\}$ be the verifier. 
For a prompt $x$, define the student's pass probability as
\begin{equation}
p_\theta(x)
=
\Pr_{y\sim \pi_\theta(\cdot\mid x)}
\big[V(y;x)=1\big].
\end{equation}
With rollout width $K\geq 2$, a valid natural preference pair exists when at least one rollout is correct and at least one rollout is incorrect. 
Thus the probability that $x$ yields a valid pair is
\begin{equation}
u_{\theta,K}(x)
=
1 - p_\theta(x)^K - \big(1-p_\theta(x)\big)^K .
\label{eq:valid-pair-prob}
\end{equation}

\paragraph{Proposition A.1}
For any $K\geq 2$, the function
\begin{equation}
u_K(p)=1-p^K-(1-p)^K,\quad p\in[0,1],\quad
\end{equation}
is uniquely maximized at $p=1/2$, with value
\begin{equation}
u_K(1/2)=1-2^{1-K}.
\end{equation}
Moreover, $u_K(0)=u_K(1)=0$.

\paragraph{Proof}
The derivative is
\begin{equation}
u_K'(p)
=
K\big[(1-p)^{K-1}-p^{K-1}\big],
\end{equation}
which vanishes only at $p=1/2$. 
The second derivative is
\begin{equation*}
u_K''(p)
=
-K(K-1)\big[p^{K-2}+(1-p)^{K-2}\big] < 0
\end{equation*}
for $p\in(0,1)$ and $K\geq 2$, so $u_K$ is strictly concave and has a unique maximum at $p=1/2$. 
The boundary values follow directly. 
\hfill $\square$

This proposition formalizes why all-correct and all-wrong prompts are inefficient: both regimes have low probability of producing usable preference pairs. 
Prompts near the competence boundary, where $p_\theta(x)$ is intermediate, provide denser supervision under the same rollout budget.

For a practice distribution $\mu$, define the expected valid-pair yield as
\begin{equation}
J_{\theta,K}(\mu)
=
\mathbb{E}_{x\sim\mu}
\big[u_{\theta,K}(x)\big].
\label{eq:yield-objective}
\end{equation}
If $M$ prompts are sampled in an iteration, the expected number of effective DPO training instances is $M J_{\theta,K}(\mu)$.

\subsection{Ideal KL-Regularized Practice Reweighting}
\label{app:kl-reweighting}

The previous result suggests that an ideal practice distribution should upweight prompts with large $u_{\theta,K}(x)$. 
Let $\mu_{\mathrm{ref}}$ be the distribution induced by a metadata-only teacher generator. 
Consider the KL-regularized update ($\eta>0$)
\begin{equation}
\label{eq:kl-objective}
\begin{split}
\mu_\eta
\in
\arg\max_{\mu\ll \mu_{\mathrm{ref}}}
\Big\{
&\mathbb{E}_{x\sim \mu}
\big[u_{\theta,K}(x)\big] \\
&-
\tfrac{1}{\eta}
D_{\mathrm{KL}}
\big(\mu \,\|\, \mu_{\mathrm{ref}}\big)
\Big\}.
\end{split}
\end{equation}

\paragraph{Proposition A.2}
The unique optimizer of \eqref{eq:kl-objective} is
\begin{align}
\mu_\eta(x)
&=
\frac{
\mu_{\mathrm{ref}}(x)
\exp\big(\eta u_{\theta,K}(x)\big)
}{
Z_\eta
}, \label{eq:exp-tilt} \\
Z_\eta
&=
\mathbb{E}_{x\sim\mu_{\mathrm{ref}}}
\left[
\exp\big(\eta u_{\theta,K}(x)\big)
\right]. \notag
\end{align}
\paragraph{Proof}
By the calculus of variations, the stationarity condition of Eq.~\eqref{eq:kl-objective} gives
\begin{equation}
u_{\theta,K}(x)
-
\frac{1}{\eta}
\left(
\log\frac{\mu(x)}{\mu_{\mathrm{ref}}(x)}+1
\right)
+
c
=0 ,
\end{equation}
where $c$ enforces normalization. 
Rearranging yields
\begin{equation}
\mu(x)
\propto
\mu_{\mathrm{ref}}(x)
\exp\big(\eta u_{\theta,K}(x)\big),
\end{equation}
which gives Eq.~\eqref{eq:exp-tilt}. 
\hfill $\square$

Eq.~\eqref{eq:exp-tilt} is an idealized exponential reweighting of the practice distribution toward high-yield prompts. 
DIAG does not execute this update exactly, since estimating $u_{\theta,K}(x)$ for a large candidate space would be expensive. 
Instead, it approximates this reweighting using observable signals from the previous iteration: topic-level valid-pair statistics and local mistake contexts.

\subsection{Mistake-Conditioned Generation as an Amortized Approximation}
\label{app:mistake-approx}

Let $z$ denote prompt metadata such as topic and difficulty, and let $\nu_t(z)$ be the scheduler over metadata at iteration $t$. 
A metadata-only generator induces a kernel $T_0(x\mid z)$, while the mistake-conditioned teacher induces
\begin{equation}
T_\phi(x\mid z,m),
\end{equation}
where $m$ contains a failed question and a wrong student trace. 
In DIAG, mistake contexts are drawn from recent prompts that produced valid natural preference pairs. 
These prompts are empirical boundary cases: the student has both successful and failed rollouts on them. 
Degenerate all-correct or all-wrong prompts are counted in yield statistics and affect pacing, but are not used as exploitation anchors.

Let $q_t(m\mid z)$ be the empirical distribution over mistake contexts for topic $z$. 
The metadata-only and DIAG-induced practice distributions are
\begin{align}
\mu_0(x)
&=
\mathbb{E}_{z\sim\nu_t}
\big[T_0(x\mid z)\big],
\\
\mu_{\mathrm{DIAG}}(x)
&=
\mathbb{E}_{z\sim\nu_t}
\mathbb{E}_{m\sim q_t(\cdot\mid z)}
\big[T_\phi(x\mid z,m)\big].
\end{align}
For each metadata slice $z$, define the local yield gain
\begin{equation}
\begin{split}
\Delta_z
={} &
\mathbb{E}_{m\sim q_t(\cdot\mid z)}
\mathbb{E}_{x\sim T_\phi(\cdot\mid z,m)}
\big[u_{\theta,K}(x)\big] \\
& -
\mathbb{E}_{x\sim T_0(\cdot\mid z)}
\big[u_{\theta,K}(x)\big].
\end{split}
\label{eq:local-yield-improvement}
\end{equation}
Then, by linearity of expectation,
\begin{equation}
J_{\theta,K}(\mu_{\mathrm{DIAG}})
-
J_{\theta,K}(\mu_0)
=
\mathbb{E}_{z\sim\nu_t}
\big[\Delta_z\big].
\label{eq:yield-decomposition}
\end{equation}
Thus, DIAG improves expected valid-pair yield whenever mistake-conditioned generation produces less degenerate follow-up prompts than metadata-only generation on average. 
This condition is not assumed as a theorem about all teachers; rather, it is the empirical property evaluated through the yield and pass-rate analyses in the main text.

\subsection{Empirical-Bayes Stabilization}
\label{app:eb-stability}

DIAG uses topic-level statistics to allocate the exploitation budget. 
For topic $z$, let $N_z$ be the number of attempted prompts and $C_z$ the number that yielded valid natural pairs. 
The raw rate $C_z/N_z$ can be noisy when $N_z$ is small. 
The Empirical-Bayes score is
\begin{equation}
v_z
=
\frac{C_z+\kappa \bar p}{N_z+\kappa}
=
\frac{N_z}{N_z+\kappa}\cdot \frac{C_z}{N_z}
+
\frac{\kappa}{N_z+\kappa}\cdot \bar p,
\label{eq:eb-convex-combination}
\end{equation}
where $\bar p$ is the global valid-pair rate and $\kappa$ is the prior strength. 
Thus $v_z$ shrinks low-count topic estimates toward the global mean, preventing the scheduler from overreacting to a few lucky samples. 
As $N_z$ grows, the topic-specific empirical rate dominates.

\subsection{Connection to DPO Curvature}
\label{app:dpo-curvature}

The signal-yield view is also consistent with the local geometry of the DPO loss. 
For a pair $(x,y_w,y_l)$, let
\begin{gather}
s_\theta = h_\theta(x,y_w) - h_\theta(x,y_l), \label{eq:s_theta} \\
\ell(s_\theta) = -\log\sigma(\beta s_\theta). \label{eq:ell_s}
\end{gather}
With respect to $a=\beta s_\theta$, the curvature is
\begin{equation}
\frac{\partial^2 \ell}{\partial a^2}
=
\sigma(a)\big(1-\sigma(a)\big),
\end{equation}
which is maximized at $a=0$. 
Hence, among valid preference pairs, updates are locally most sensitive when the model is uncertain between the preferred and dispreferred responses. 
This does not equate prompt-level pass probability with pairwise DPO uncertainty, but it supports the same design principle: practice near the student's moving boundary tends to produce more informative supervision.

\section{Complexity Analysis}
\label{sec:complexity}

We analyze computational cost through the lens of \emph{effective supervision}.
Let $C_{\text{roll}}$ be the unit cost of one student rollout and $C_{\text{teach}}$ the unit cost of one teacher generation.
At iteration $t$, we use a practice set $P_t$ of size $|P_t|=M$ and sample $K$ rollouts per question, so the student-side cost per iteration is
\begin{equation}
\mathcal{C}^{\text{stu}}_t = MK\,C_{\text{roll}}.
\end{equation}
We focus on verifiable preference learning where a question $x$ yields a usable DPO pair iff it contains both correct and incorrect rollouts (Eq.~\eqref{eq:valid_pair}).
Recall the {effective (valid-pair) yield} is defined as
\begin{equation}
\rho_t \;\triangleq\; \frac{1}{M}\sum_{x\in P_t}\mathbb{I}_{\text{valid}}(x),
\end{equation}
following Eq.~\eqref{eq:yield}. 
The expected number of effective training pairs in iteration $t$ is
\begin{equation}
\mathbb{E}[|\mathcal{D}_t|] \;=\; M\rho_t.
\end{equation}

\paragraph{Compute-matched teacher budget.}
In our experiments, \textsc{Static Gen} and \method are matched by the same total number of teacher-generated questions.
Concretely, over $T$ iterations both methods use $TM$ teacher-generated questions, implying identical total teacher cost:
\begin{equation}
\mathcal{C}^{\text{teach}}_{\text{total}} \;=\; TM\,C_{\text{teach}}.
\end{equation}
The difference is \emph{not} the amount of teacher compute, but \emph{how} this fixed teacher budget is spent:
\textsc{Static Gen} generates a fixed pool using only (topic, difficulty), while \method allocates the same number of generations online and conditions on the student's recent failures to reshape the practice distribution.

\paragraph{Cost per effective pair (effective gradient cost).}
Since the student rollout budget per iteration is fixed, the key efficiency question is: \emph{how many of these rollouts produce gradient-bearing preference pairs?}
Define the per-iteration compute cost per effective pair as   
\begin{equation}
\label{eq:eff_cost}
\begin{split}
\mathcal{E}_t 
&\;\triangleq\;
\frac{\mathcal{C}^{\text{stu}}_t + \mathcal{C}^{\text{teach}}_t}{\mathbb{E}[|\mathcal{D}_t|]}
\;=\;
\frac{MK\,C_{\text{roll}} + \mathcal{C}^{\text{teach}}_t}{M\rho_t} \\
&\;=\;
\frac{K\,C_{\text{roll}}}{\rho_t} + \frac{\mathcal{C}^{\text{teach}}_t}{M\rho_t}.
\end{split}
\end{equation}
Under compute matching, $\mathcal{C}^{\text{teach}}_t$ is the same order for both methods (and the total teacher budget is identical), therefore the dominant factor is the yield $\rho_t$.
When a static pool (or practice-agnostic generation) degenerates into all-correct or all-wrong regimes, $\rho_t$ decreases and $\mathcal{E}_t$ increases, meaning a growing fraction of rollouts are wasted with no contrastive supervision.

\paragraph{\method improves effective sample efficiency.}
By explicitly optimizing the practice distribution to stay near the student's moving decision boundary, \method sustains a substantially higher $\rho_t$ across iterations (Sec.~\ref{sec:analysis}).
Plugging this into Eq.~\eqref{eq:eff_cost} yields a strictly lower compute-per-effective-pair:
\begin{equation}
\begin{split}
&\rho_t^{\textsc{\method}} > \rho_t^{\textsc{Static Gen}} \\
&\qquad\Longrightarrow\quad
\mathcal{E}_t^{\textsc{\method}} < \mathcal{E}_t^{\textsc{Static Gen}}.
\end{split}
\end{equation}
Equivalently, for the same fixed rollout budget $MK$, \method produces more effective preference pairs $M\rho_t$ and therefore more gradient signal per unit compute.
This explains why \method achieves higher final performance without increasing either student rollout compute or the total teacher-generation budget.

\section{Empirical Bayes Derivation and Implementation Details}
\label{app:eb}

This appendix provides a brief derivation of the Empirical Bayes (EB) topic score used in Phase~I and summarizes the corresponding implementation choices.

\paragraph{Beta--Binomial model.}
For each topic $z$, let $p_z$ denote the latent probability that a sampled question from topic $z$ yields a valid \emph{natural} preference pair (Eq.~\ref{eq:valid_pair}). 
Given per-topic counts $(C_z,N_z)$ (Eq.~\ref{eq:NzCz_def}), we model
\begin{gather}
p_z \sim \mathrm{Beta}(\alpha_0,\beta_0), \notag \\
C_z \mid (p_z,N_z) \sim \mathrm{Binomial}(N_z,p_z).
\end{gather}
By conjugacy, the posterior is
\begin{equation}
\begin{split}
p_z \mid (C_z,N_z) \sim \mathrm{Beta}\big(&\alpha_0 + C_z,\\
&\beta_0 + N_z - C_z\big),
\end{split}
\end{equation}
and the posterior mean yields the EB score
\begin{equation}
\label{eq:eb_appendix_mean}
v_z \triangleq \mathbb{E}[p_z \mid C_z,N_z] 
= \frac{C_z + \alpha_0}{N_z + \alpha_0 + \beta_0}.
\end{equation}

\paragraph{Empirical Bayes prior from global statistics.}
We set the Beta prior using the global valid-pair rate $\bar{p}$ and a smoothing strength $\kappa$ (prior sample size):
\begin{equation}
\alpha_0 \triangleq \kappa \bar{p}, \qquad \beta_0 \triangleq \kappa(1-\bar{p}).
\end{equation}
Substituting into Eq.~\ref{eq:eb_appendix_mean} gives the shrinkage estimator used in the main text:
\begin{equation}
\label{eq:eb_appendix_shrinkage}
v_z = \frac{C_z + \kappa \bar{p}}{N_z + \kappa}.
\end{equation}
When $N_z \ll \kappa$, the score shrinks towards $\bar{p}$, preventing over-reaction to low-count noise; as $N_z$ grows, the empirical rate $C_z/N_z$ dominates.

\paragraph{Implementation details.}
We compute $\bar{p}$ as the running global average of $\mathbb{I}_{\text{valid}}(x)$ over all attempted questions up to the current iteration (equivalently, $\bar{p}=\sum_z C_z / \sum_z N_z$ using the same statistics as Eq.~\ref{eq:NzCz_def}). 
We fix $\kappa$ as a small constant in all experiments, which controls the strength of shrinkage in the early, noisy regime.

\section{Prompts}

\paragraph{Prompt Templates.}
We present the prompt templates employed for the Diagnose-and-Generate loop in Figure~\ref{fig: Prompt Teacher} and Figure~\ref{fig: Prompt Student}. 
For the teacher model, the instruction explicitly incorporates the \textit{Mistake Context}, comprising the failed question and the erroneous solution trace, instructing the model to synthesize a variant that probes the same underlying misconception while altering surface features.
Regarding student rollouts, we employ a standard format to elicit reasoning. To harmonize the reasoning structure across heterogeneous model architectures, we adopted an {adaptive decoding strategy}: for models requiring explicit structural guidance, the generation was primed with a sequential prefix (e.g., ``{Step 1:}'' \citep{lai2024step}) to initiate the deduction process; for models with strong intrinsic instruction adherence, standard generation was utilized.
\begin{figure*}[htpb]
\begin{tcolorbox}[colback=black!5!white, colframe=black!75!black, 
     title=Prompt for Student Model Answer Generation]
\texttt{<|im\_start|>system} \\
Please reason step by step, and put your final answer within \textbackslash boxed\{\}. \\
\texttt{<|im\_end|>} \\
\texttt{<|im\_start|>user} \\
\{problem\} Let's think step by step and output the final answer within \textbackslash boxed\{\}. \\
\texttt{<|im\_end|>} \\
\texttt{<|im\_start|>assistant}
\end{tcolorbox}
\begin{tcolorbox}[colback=black!5!white, colframe=black!75!black, 
      title=Prompt for Teacher Model Answer Generation]
\texttt{<|im\_start|>system} \\
You are a helpful assistant. \\
\texttt{<|im\_end|>} \\
\texttt{<|im\_start|>user} \\
\{question\} Let's think step by step and output the final answer within \textbackslash boxed\{\}. \\
\texttt{<|im\_end|>} \\
\texttt{<|im\_start|>assistant}
\end{tcolorbox}
\caption{Prompt for Student Model and Teacher Model Answer Generation.}\label{fig: Prompt Student}
\end{figure*}

\begin{figure*}[htpb]
\begin{tcolorbox}[colback=black!5!white, colframe=black!75!black, 
     title=Prompt for Teacher Model Question Generation]
You are an expert math curriculum designer and a master question creator. Your task is to generate high-quality, targeted training questions for a language model ("StudentLM"). Every question must be rigorous, original, aligned with the requested topic and difficulty, and directly target the student's weaknesses. You need to ensure that the answer to the generated question is a number or a formula, rather than a sentence or a proof. \\\\
\#\#\# 1. Context \& Goal \#\#\# \\
- General Topic: \{topic\} \\
- Difficulty Level: \{difficulty\} \\\\
\#\#\# 2. Student Model Error Samples \#\#\# \\
\{Error Sample: [Question (with Gold Solution), Student Failure]\} \\\\
\#\#\# 3. Additional Guidance \#\#\# \\
\{guidance\} \\\\
\#\#\# 4. Output Format \#\#\# \\
First, analyze the student's error in a \texttt{<guidance>} block: identify the core misconception 
in the student's reasoning. Then, design a new problem that specifically forces the student to confront this misconception. Output in this exact structure: \\\\
\texttt{<guidance>} \\
...brief analysis of the student's misconception... \\
\texttt{</guidance>} \\\\
\texttt{<question>} \\
...problem statement... \\
\texttt{</question>} \\\\
Do NOT include any solutions, answers, or extra commentary outside these two blocks.
\end{tcolorbox}
\caption{Prompt for Teacher Model Question Generation.}\label{fig: Prompt Teacher}
\end{figure*}

 \section{Implementation Details}\label{app:implementation}                    
                                                                                
  We provide the detailed hyperparameters and system configurations used in our 
  experiments to ensure reproducibility.                                        
\subsection{Training Configuration}                                           
All student models (Qwen2.5-Math-7B, Qwen3-8B-Base, Llama-3.1-8B) were trained using the DPO objective with the AdamW optimizer, a cosine learning rate schedule with peak learning rate $5\times10^{-7}$, no weight decay, global batch size 128 (gradient accumulation 16), and 2 epochs per iteration. For DPO, we set the KL penalty $\beta=0.1$, use sigmoid loss with label smoothing $0.1$, and freeze the previous iteration's checkpoint as the reference policy. Sequences are truncated at 4096 tokens (prompts at 1000), and training uses bfloat16 precision with gradient checkpointing enabled.

Regarding data attributes, we employ a hierarchical taxonomy annotated by Gemini-2.5~\citep{comanici2025gemini}. This includes five primary domains: \textit{Algebra, Geometry, Calculus, Discrete Mathematics, and Probability}. Each contains several fine-grained sub-topics (e.g., \textit{Polynomials}, \textit{Modular Arithmetic}), along with four difficulty levels ranging from \textit{Elementary} to \textit{Expert}. For mistake-conditioned generation, failure traces are sampled from the mistake buffer $\mathcal{H}_{t-1}$ via an uncertainty-weighted probability proportional to $1-\max(\hat{p}, 1-\hat{p})$ to prioritize boundary cases, which is standard in \cite{settles2009active}. The pacing factor $\lambda$ in \eqref{eq:pacing_alpha} is set to $0.5$ by default.           

  \subsection{Generation \& Rollout Configuration}                              
  The quality of the feedback loop relies heavily on the decoding strategies    
  employed during the rollout phase.                                            
  \begin{itemize}[leftmargin=*,nosep]
      \item \textbf{Student Rollouts:} For collecting preference data, we sample
   $K=8$ responses per prompt
   with temperature $T=1.0$, top-$p=0.99$, and top-$k=50$ to ensure diversity in
   reasoning paths.                                                              
      \item \textbf{Teacher Generation:} For the targeted curriculum synthesis, 
  the teacher model uses temperature $T=1.0$ with top-$p=0.99$ for question     
  generation. For answer generation, we use greedy decoding ($T=0$) to maximize correctness. 
  \item \textbf{Verifier:} The verifier adopts rule-based evaluation \citep{zhangonline} to determine correctness.
                                                                   
  \end{itemize}                                                                 
\subsection{Cross-Task Evaluation Implementation Details}
\label{appsubsec:implementation}

To ensure reproducibility, we adopt standard evaluation protocols for all benchmarks.
For \textbf{BBH} and \textbf{MMLU}, we utilize deterministic greedy decoding ($T=0$) with a single rollout.
For \textbf{TruthfulQA}, we follow the standard MC1 and MC2 validation setup, calculating scores based on the log-probabilities of candidate answers.
For \textbf{HumanEval}, we employ a sampling-based approach: we generate $k=20$ solutions per problem with a temperature of $T=0.1$ and report the average pass rate.

\section{Case Studies}

\subsection{Ignoring Interior Singularities in Improper Integrals}
\label{subsec:case_improper_integral}

A persistent failure mode among small models is mechanically applying the Fundamental Theorem of Calculus without checking whether the integrand is continuous on the integration interval. In the seed question below, the student is asked to evaluate $\int_{-1}^{1}1/x^{2}\,dx$. The integrand has a singularity at $x=0$, making the integral divergent, but the student simply computes the antiderivative $-1/x$, substitutes the endpoints, and reports $-2$, which is a negative value for a non-negative integrand, an inconsistency that goes entirely unnoticed.

The follow-up shifts the singularity from $x=0$ to $x=2\in(1,4)$ by asking the student to evaluate $\int_{1}^{4}1/(x-2)^{2}\,dx$. This defeats the shallow heuristic ``watch out when $x=0$ is in the interval'' while preserving the same underlying misconception. The na\"ive FTC evaluation now yields $-3/2$, again negative for a strictly positive integrand, so even a minimal sign-check would catch the error. Students who have internalized the continuity prerequisite immediately output $0$ (divergent); those who have not will consistently report $-3/2$, producing a clean binary split in rollout outcomes.

\subsection{Base-Rate Neglect in Bayesian Inference}
\label{subsec:case_bayes}

The second case targets a different kind of oversight: ignoring prior probabilities when computing a posterior. Two machines have different production shares (60\% vs.\ 40\%) and defect rates (5\% vs.\ 3\%). The student is asked for $P(A\mid\text{defective})$ and normalizes the likelihoods directly, arriving at $5/(5+3)=5/8$. The correct answer is $5/7$, obtained by weighting each likelihood by its prior. Notably, the student's self-check ($5/8+3/8=1$) is tautological, since any normalization sums to one, yet it gives a false sense of correctness.

To expose this misconception more sharply, the follow-up keeps the same defect rates but pushes the priors to an extreme: Machine~A now produces only $1\%$ of items while Machine~B produces $99\%$. The likelihood-only shortcut yields the same $5/8\approx62.5\%$, but the correct Bayesian posterior drops to $0.0005/0.0302\approx1.66\%$, which is off by a factor of nearly $40\times$. The dramatic gap ensures that any model still ignoring the prior will produce an answer that is obviously wrong, rather than merely slightly off.

\begin{figure*}[htpb]
\centering

\begin{tcolorbox}[colback=black!5!white, colframe=black!75!black,
     title=Student Answer (Incorrect) -- Improper Integral]
\textbf{Step 1:} Find the antiderivative of $\frac{1}{x^2}$. We know that $\frac{1}{x^2} = x^{-2}$, so the antiderivative is $\frac{x^{-1}}{-1} = -\frac{1}{x}$.

\textbf{Step 2:} Apply the Fundamental Theorem of Calculus. That is $\int_{-1}^{1}\frac{1}{x^{2}}\,dx = \left[-\frac{1}{x}\right]_{-1}^{1}$.

\textbf{Step 3:} Evaluate at the bounds. Upper bound: $-\frac{1}{1} = -1$. Lower bound: $-\frac{1}{-1} = 1$.

\textbf{Step 4:} Compute the result.
$$\left[-\frac{1}{x}\right]_{-1}^{1} = (-1) - (1) = -2.$$
The answer is $\boxed{-2}$.
\end{tcolorbox}
\vspace{-2mm}
\begin{tcolorbox}[colback=black!5!white, colframe=black!75!black,
     title=Teacher-Generated Follow-up (Targeted) -- Improper Integral]
Evaluate the integral $\displaystyle\int_{1}^{4}\frac{1}{(x-2)^{2}}\,dx$.
If the integral diverges, enter~$0$.
\medskip

\noindent\textbf{Correct Answer:} $\boxed{0}$ \;

\noindent\textit{Trap answer via direct FTC:}
$\bigl[-\tfrac{1}{x-2}\bigr]_{1}^{4}=-\tfrac{1}{2}-1=-\tfrac{3}{2}$.
\end{tcolorbox}

\vspace{3mm}

\begin{tcolorbox}[colback=black!5!white, colframe=black!75!black,
     title=Student Answer (Incorrect) -- Bayesian Inference]
\textbf{Step 1:} Identify the given information.
Machine A: defect rate = 5\%. Machine B: defect rate = 3\%. We are told a defective item was found, and we need $P(A \mid \text{defective})$.

\textbf{Step 2:} Since the item is already known to be defective, we need to determine which machine is more likely to have produced it. We can compare the defect rates directly.
The probability of being defective given Machine A is 5\%, and given Machine B is 3\%.

\textbf{Step 3:} Normalize to get the posterior.
$$P(A \mid \text{defective}) = \frac{P(\text{defective} \mid A)}{P(\text{defective} \mid A) + P(\text{defective} \mid B)} = \frac{0.05}{0.05 + 0.03} = \frac{5}{8}.$$

\textbf{Step 4:} Verify. $\frac{5}{8} + \frac{3}{8} = 1$. The probabilities sum to 1, so the answer is consistent.
The answer is $\boxed{\dfrac{5}{8}}$.
\end{tcolorbox}
\vspace{-2mm}
\begin{tcolorbox}[colback=black!5!white, colframe=black!75!black,
     title=Teacher-Generated Follow-up (Targeted) -- Bayesian Inference]
A factory has two machines. Machine~A produces $1\%$ of all items and has a $5\%$ defect rate.
Machine~B produces $99\%$ of all items and has a $3\%$ defect rate.
If a randomly selected item is found to be defective, what is the probability it was produced by Machine~A?
\medskip

\noindent\textbf{Correct Answer:} $\frac{0.0005}{0.0302}\approx1.66\%$.

\noindent\textit{Trap answer via likelihood-only normalization:}
$\frac{5}{5+3}=\frac{5}{8}=62.5\%$.
\end{tcolorbox}

\caption{\textbf{Case Studies.}
\textit{Top pair (Improper Integral):} The student applies FTC across a singularity at $x=0$, arriving at $-2$; the follow-up shifts the singularity to $x=2$, where the trap value $-3/2$ is negative for a strictly positive integrand.
\textit{Bottom pair (Bayesian Inference):} The student normalizes likelihoods without prior weighting ($5/8$ instead of $5/7$); the follow-up pushes priors to $1\%$ vs.\ $99\%$, where the same shortcut gives $62.5\%$ but the correct posterior is $\approx1.66\%$.}
\label{fig:case_studies_combined}
\end{figure*}
\end{document}